# Proposing Plausible Answers for Open-ended Visual Question Answering

**Omid Bakhshandeh**
University of Rochester
omidb@cs.rochester.edu

**Trung Bui**
Adobe Research
bui@adobe.com

**Zhe Lin**
Adobe Research
zlin@adobe.com

**Walter Chang**
Adobe Research
wachang@adobe.com

## Abstract

Answering open-ended questions is an essential capability for any intelligent agent. One of the most interesting recent open-ended question answering challenges is Visual Question Answering (VQA) which attempts to evaluate a system's visual understanding through its answers to natural language questions about images. There exist many approaches to VQA, the majority of which do not exhibit deeper semantic understanding of the candidate answers they produce. We study the importance of generating plausible answers to a given question by introducing the novel task of 'Answer Proposal': for a given open-ended question, a system should generate a ranked list of candidate answers informed by the semantics of the question. We experiment with various models including a neural generative model as well as a semantic graph matching one. We provide both intrinsic and extrinsic evaluations for the task of Answer Proposal, showing that our best model learns to propose plausible answers with a high recall and performs competitively with some other solutions to VQA.

## 1 Introduction

With the recent progress made in AI, there is a renewed interest in building AI systems which are capable of reasoning in addition to perception. As humans, we often use our commonsense reasoning for interpreting complex visual and auditory input. Question Answering (QA) is a crucial ability for any intelligent system and requires many degrees of complex reasoning. Imagine that you are blindfolded and asked 'What is parked next to that tree?'. Although you cannot see the referent objects, you can still propose a set of plausible answers using your common sense. Your set of proposed answers is most probably within {car, bike, bus, motorcycle, scooter} or other objects that can be parked. As humans, given the semantic interpretation of a question, we have a set of default presumptions about the semantic type of the plausible answer. Each question seems to reflect some semantic 'frame' which then naturally activates certain 'slots' that can only be instantiated with certain types of entities. Hence, one could link the surface task of 'finding plausible answers to a question' to a deeper theory of structured commonsense knowledge. As Minsky (1974) points out, a question also includes suggestions and recommendations about its set of answers (assignment proposal). Minsky notes that " 'default' assignments become the simplest special cases of recommendations, ... one has a hierarchy in which such proposals depend on features of the situation".

In this paper we focus on developing the capability to propose relevant and plausible answers to a given question, which is a key intelligent behavior that an AI system should demonstrate. We introduce the novel task of *'Answer Proposal'*, in which a system seeks to generate a ranked list of meaningful candidate answers associated with the semantic features of a given question. Having a prior commonsense knowledge about the scope of the plausible answers can not only narrow down the search space for the final prediction, but also, in case of an incorrect prediction, help the system appear more intelligent from the user-experience point of view.

A great framework for showcasing the potential of Answer Proposal is multimodal QA. Visual Question Answering (VQA) (Antol et al., 2015; Ren et al., 2015) is one of the most interesting multimodal

| 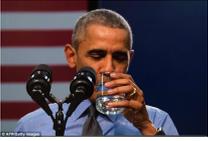 | 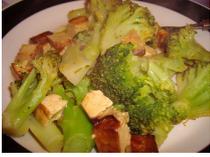 | 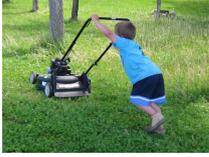 | 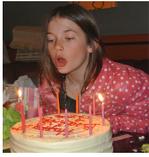 |
|---|---|---|---|
| Q. What is he drinking out of? | What is the dominant vegetable? | What is he mowing? | What is she blowing out? |
| A. {wine, beer, water, 1, 4} | {broccoli, green, 1, cheese, carrot} | {fire hydrant, frisbee, bat, kite, 2 } | {sprinkles, cheese, candles, ketchup, tomato} |

Table 1: Example images along with their corresponding questions (Q) and the set of top answers (A) generated by SOTA-QA system. Given the semantics of the question, the answers in red are not plausible commonsense answers disregarding what the image is.

QA tasks for evaluating visual understanding of images through questions posed in natural language. This task is set up as follows: given an image and a natural language question about it, a system should provide an accurate natural language answer. The answer is either selected from a list of choices (for a multiple-choice question), or is generated (for an open-ended question), where the open-ended task is more challenging than the multiple-choice one. In both cases, VQA is automatically evaluated given that many open-ended answers contain only a few words.

There have been many recent approaches for tackling the task of VQA. The state-of-the-art open-ended systems mainly train a multi-class classifier which uses the space of all possible answers for classification. If we look under the hood of such systems, it is evident that they do not have any deep understanding of whether their top answers are even plausible candidates for the given question. For example, consider the example images along with their corresponding questions in Table 1. The Answer row of this table shows the top answers from one of the state-of-the-art systems (hereinafter, SOTA-QA)[1] (Antol et al., 2015). Looking at these top answers, it is clear that this system does not exhibit basic understanding of e.g., the kind of things that you can drink out of, or the kind of things that can be blown out.

In this paper we mainly leverage the power of Answer Proposal for proposing plausible answers for open-ended questions in the context of the VQA problem. The contributions of this paper can be summarized as follows: (1) We introduce the task of *'Answer Proposal'* together with an intrinsic metric for stand-alone evaluation (Section 2). (2) We present various Answer Proposal models, ranging from a neural generative model to a semantic graph matching one. We tackle the task of VQA by feeding the Answer Proposal list into a deep binary classifier which determines the correctness of a proposed answer (Section 3-4). (3) We show that our best answer proposal model achieves a high recall score and generates highly plausible answer proposal sets. Furthermore, our approach for open-ended VQA is competitive with some other models and can also unveil some of the biases of multiple-choice VQA (Section 5). We hope that the results in this paper ignite interest in the community to leverage semantic understanding and commonsense knowledge for tackling VQA.

## 2 The Task of Answer Proposal

We define the task of *'Answer Proposal'* as follows.

**Definition 2.1.** *Given the question $Q$, create a list of all plausible answers, $P$, which is ranked according to their prior probabilities.*

For example, given the question $q =$'What is parked in front of the tree?', the set $p =$ {'car', 'bike', 'motorcycle'}. The objective of a system is to generate the list $P$ in a way that the actual correct answer appears higher in the ranked list. Hence, we define the intrinsic evaluation of the task as follows.

**Definition 2.2.** *Given a list of $M$ triplets of questions, answers, and the plausible answer proposal list, such as $(q_i, a_i, p_i)$, we define Recall@N as:*

---
[1]Demo available through http://cloudcv.org/vqa/

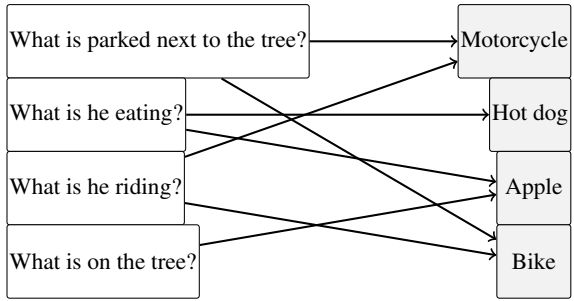

Figure 1: Example many-to-many mapping between questions and the plausible answers.

$$Recall@N = \frac{\sum_{i=1 to M} I_{iN}}{M} \quad (1)$$
$$I_{iN} = 1 \; if \; a_i \in p_i[:N], \; else \; 0.$$

where $P_i[:N]$ is the answer proposal list with cutoff=$N$ and $I_{iN}$ is the success indicator variable. Recall@N evaluation metric has been used in information retrieval for evaluating the quality of the retrieved ranked lists, hence, is a great fit for the task of Answer Proposal.

The problem of finding plausible answer set given a question can be viewed as a many-to-many mapping. Figure 1 draws an example such mapping. It is important to note that learning such a mapping accurately is challenging, e.g., 'Apple' is a plausible answer for both 'What is he eating?' and 'What is on the tree?' questions despite the fact that these questions are not semantically similar. On the other hand, 'Hot dog' is an edible entity but cannot be found on top of a tree.

## 3 Approach for Tackling VQA

We extrinsically evaluate Answer Proposal through the VQA task. Our approach for tackling VQA consists of two main modules: answer proposal generator, and a simple deep binary classifier. The answer proposal module takes in the question and then generates a list of plausible proposal answers ($p$) using the semantic features of the question ($q$). Then each item from the proposal list ($p$), together with the question ($q$) and the image ($i$), gets fed into the deep binary classifier which then predicts the probability of the triple ($q, i, p$) being correct. Figure 2 shows this pipeline. In Section 4, we will introduce various models for Answer Proposal module.

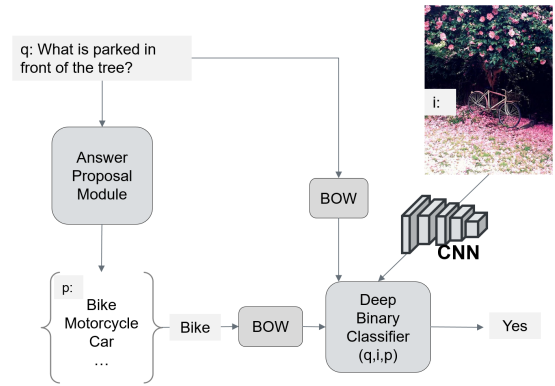

Figure 2: Answer Proposal approach for tackling the task of VQA.

**Deep Binary Classifier**

This module classifies whether or not a given triplet $(q, i, p)$ is correct. Our deep binary classifier is a simple multilayer perceptron which works as follows: it takes in the concatenated feature vectors representing each of the three inputs, and outputs a prediction of 'yes' or 'no' indicating the correctness of the triplet. Hence, our model is $\sigma(W^T F_{qip})$, where $F_{qip}$ refers to the concatenated feature vectors and $\sigma$ is the sigmoid function. Our image features are 2,048-dimensional vectors computed using the penultimate layer of the state-of-the-art convolutional neural network for image recognition, Resnet-101 (He et al., 2016). Both the question and plausible answer are represented using 300-dimensional average Word2Vec word embeddings (Mikolov et al., 2013) which is a bag-of-words (BOW) model[2]. The classifier is trained end-to-end with the objective of minimizing the binary logistic loss of the prediction by using stochastic gradient descent. The model has three layers, where each layer has 8,192 hidden units, with dropout after the first layer.

## 4 Answer Proposal Models

In this Section we present various models for generating Answer Proposal lists. We mainly devise two classes of approaches: generative and retrieval.

---
[2]We also experimented with using a Recurrent Neural Network for encoding the question, which yielded worse results (Jabri et al., 2016; Zhou et al., 2015).

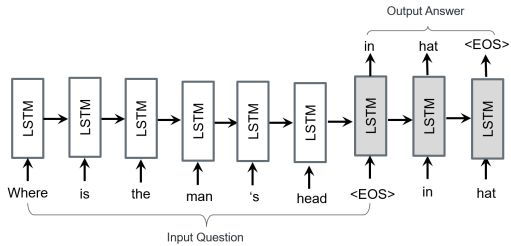

Figure 3: The generative Answer Proposal model.

### 4.1 Generative Model

The generative model is an encoder-decoder Recurrent Neural Network (RNN) architecture (Sutskever et al., 2014; Cho et al., 2014), which generates the answer proposal being conditioned on the question. The encoder RNN processes the question and the decoder[3] generates the proposed answer one token at a time until hitting the end-of-sentence (EOS) token. Every question is encoded into a state vector of size 512, which is then set as the initial recurrent state of the decoder. We tune the model parameters on the val set, where we set the number of layers to 2. The model is trained end-to-end, using Stochastic Gradient Descent with early stopping. For decoding, we use beam search with beam-width = 25. The main advantage of this model is that it can compose novel multi-word answers which have not been seen in the training set before. Figure 3 outlines this model by an example.

### 4.2 Retrieval Models

A retrieval model uses similarity metrics to retrieve train-set questions similar to the test question. There are various ways for capturing semantic similarity between a pair of questions. In this Section we present two different retrieval approaches.

**Average Word2Vec Model**

We experimented with various textual similarity metrics, among which BLEU and Word2Vec are notable[4]. BLEU is the widely used Machine Translation (MT) metric (Papineni et al., 2002) which scores a hypothesis against a gold reference by computing the geometric mean of precision scores for different n-grams. BLEU can only capture exact n-gram matches, hence, for instance, it does not account for the similarity between 'eating' and 'devouring' for comparing 'What is she eating?' and 'What is he devouring?'. We obtained the most promising results when using Word2Vec (W2V) as the similarity metric. We use Word2Vec (Mikolov et al., 2013) as a sentence-level vector representation where we average the word-level embeddings to obtain the sentence-level vector. Then, the similarity between two sentences equals the cosine similarity between their corresponding vectors. This simple model is effective in capturing generic similarity of many questions such as 'What is filled with water?' and 'What is she filling with water?'.

**Semantic Graph Matching Model**

The generative model represents the natural language question as a stream of tokens and the Word2Vec model represents it as a bag of words. As discussed in Section 1, as humans, given the rich semantic features of a question, we can infer a set of default presumptions about the semantic types of plausible answers. Our semantic graph matching model attempts to get closer to that premise by encoding the semantic structure of each question. There are various approaches for representing the structure of questions, including dependency parse trees or semantic parses. Although the dependency structure provides a lot of information regarding how the individual words relate grammatically, it does not provide much information regarding what a sentence actually means and what the ontology types of different words are[5]. This makes deep semantic parsing a more suitable choice for our task. Semantic parse graph of a sentence maps natural language input to a formal meaning representation. A broad-coverage semantic parser (Banarescu et al., 2013; Bos, 2008; Allen et al., 2008) operates at the generic natural language level, mapping surface level words into their underlying meaning representation.

---

[3] We got worse results using attention for decoder RNN.

[4] We also tried other sentence-level embedding models, such as the Skip-thoughts model (Kiros et al., 2015), all of which performed weaker than Word2Vec in capturing generic textual similarity

[5] Beyond these shortcomings, the state-of-the-art dependency parsers (e.g., CoreNLP (Manning et al., 2014)) very often fail at parsing questions altogether, mainly confusing copular and auxiliary constructions. This is due to their training corpora which often lacks questions.

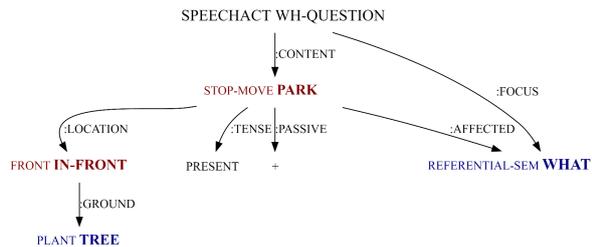

Figure 4: The semantic parse graph for the question 'What is parked in front of the tree?'.

Here we use the TRIPS[6] (Allen et al., 2008) broad-coverage semantic parser which produces the state-of-the-art logical form from natural text (Allen et al., 2008). The TRIPS logical form language is an encoding of the semantic content of a sentence that can be mapped to a formal knowledge representation. TRIPS provides a richer semantic structure than other off-the-shelf text processing systems, mainly it provides sense disambiguated deep structures augmented with semantic ontology types[7]. Figure 4 shows the TRIPS semantic parse for the question 'What is parked in front of the tree?'. In this graph representation, each node specifies a word [8] in bold along with its corresponding ontology type on its left. The edges in the graph are semantic roles[9] between the nodes. The root of each graph is an speech act node. SPEECHACT indicates the communicative function of an utterance, such as 'tell' act or an 'acknowledgment' one. The two speech act types that we are interested in are the following:

– WH-QUESTION: Indicates a 'wh' question node, which has two semantic roles: (1) a 'content' role which points to the main semantic type in the question and (2) a 'focus' role which refers to the 'wh' question word itself.

– YES/NO-QUESTION: Indicates a Yes/No question, which only has the semantic role 'content'.

As shown in Figure 4, such a semantic parse graph provides a semantically rich representation for the question.

As a retrieval model, our semantic graph matching approach aims at using the semantic parse graph representations to retrieve similar train-set parse graphs given a test parse graph. This approach consists of the following two main stages:

• **Stage 1, question type categorization**: As mentioned earlier, the semantic features of the questions play a major role in our understanding of the plausible answers. The question word, represented by the SPEECHACT node, identifies the subcategory of answer types. Mainly, the only plausible answer for YES/NO-QUESTIONs are {yes, no}[10], or the answer to WH-QUESTION of type 'How many' is always a number. At this stage, given a test question, we filter a part of the train-set which share the same question type as with the test set.

• **Stage 2, semantic graph matching**: At this stage, given a set of train-set graphs which share the same question type category from stage 1, we semantically match the test graph with the train-set graphs. Consider the two questions 'What is she eating?' and 'What is he consuming in the kitchen?'. The graph of these sentences would not exactly match, however, they are indeed similar and share the same plausible answer set. This brings up the idea of implementing different graph mutation patterns to mutate the test graph through performing a few actions, each with a different priority. The mutations are a combination of the following two actions: (1) replacing a node with its ontology type (one or more levels up), or (2) deleting a node[11]. At the end, this stage generates a ranked list of plausible answers, where a matching between a train-set graph and less mutated test graph appears higher in the list.

## 5 Experiments

In this Section we summarize our experiments on intrinsic and extrinsic evaluation of various Answer Proposal models. For all the experiments, we use the COCO trainval2014 dataset with the same train

---

[6] http://trips.ihmc.us/parser/cgi/parse

[7] More importantly, TRIPS parser was originally developed as a part of a conversational assistant, tailored to parsing natural language questions.

[8] What is not shown in this graph is that words are also sense disambiguated according to WordNet (Miller, 1995) senses.

[9] For the full list of semantic roles in TRIPS parser please refer to http://trips.ihmc.us/parser/LFDocumentation.pdf.

[10] Our semantic graph matching model does not need to proceed to further stages for the YES/NO-QUESTIONS as the ultimate plausible answer set is {yes, no}.

[11] Linguistic knowledge about core semantic role of verbs enables us to make an informed decision about deleting the nodes.

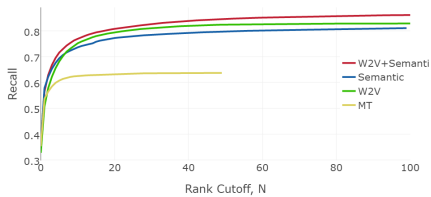

(a) Against the *majority* human answer.

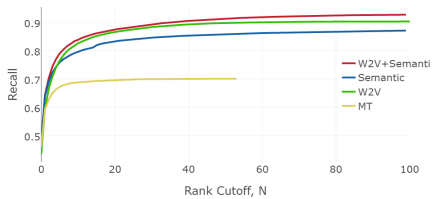

(b) Against *any* human answer.

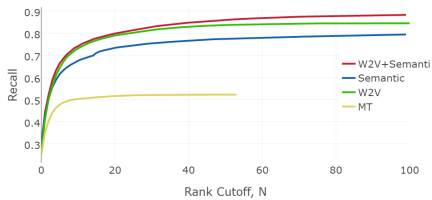

(c) Against *any* human answer, only wh-questions.

Figure 5: Intrinsic evaluation according to the Recall@N scores.

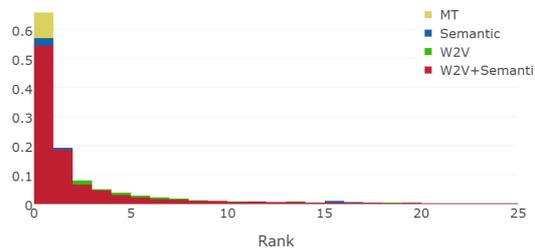

Figure 6: The rank distribution of the correct answers within the answer proposal lists.

and val set split as with Zhou et al. (2015), containing 339,482 training and 30,377 test instances. We use the test2015-standard blind set as the test set. In addition to the models described in Section 4, we include the model *W2V+Sem* in the experiments. Given the ranked proposal lists from the *Semantic* and the *W2V* model, the *W2V+Sem* model simply generates an aggregated ranked list by alternating between the ranked items of the two lists.

### 5.1 Intrinsic Evaluation

We intrinsically evaluate various models according to their Recall@N score. Each open-ended VQA question come with a gold answer list of size 10. Here we compute Recall@N scores according to two measures of correctness: (1) it exactly matches the *majority* answer, (2) it matches *any* of the answers in the answer list. Given that the blind test set does not provide the answers, we perform the intrinsic evaluation on the trainval2014-val set.

Figure 5a shows the Recall@N scores evaluated according the *majority* answer and Figure 5b shows the evaluation against *any* of the answers. These results show that our best model perform very well (with the highest recall of 87%@100 according to *majority* and 93%@100 according to *any*) on generating relevant answers to a given question. The model traces become almost constant after N=100. As this graph shows, the *W2V+Sem* model achieves the highest Recall and the *Semantic* approach and *W2V* are competitively close. The high recall of *W2V* retrieval model shows the effectiveness of word embeddings for measuring the similarity among VQA questions. This further suggests that there are many structurally similar questions in the VQA dataset which makes capturing word order in meaning representation less crucial. The MT model comes short in generating long hypotheses list and is the weakest performing system. It is interesting to see that our best models obtain a high recall on wh-questions (Figure 5c) as well.

Although the MT model often fails at including the correct answer within its proposal list, whenever the correct answer is included in the proposal (i.e., there is a *hit*), it is at the very top of the list. Figure 6 visualizes the rank distribution of all the *hits* across all the models, where the MT model shows to have more than 60% of its answers at rank 1. Moreover, as expected, the MT model is capable of generating novel answers (e.g., 'wooden bottle'), however, it suffers from generating generic and commonplace answers (e.g., '0 feet', as an answer for a 'how long' question). As a result, we did not include the MT

| | Q. | What is the bear on the left wearing? 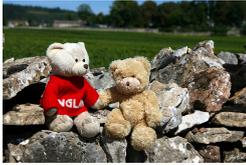 | How long is this animal's neck? 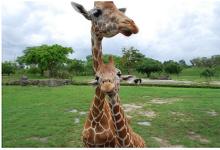 | What are the zebras drinking from? 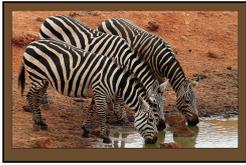 |
|---|---|---|---|---|
| MT | | { 'overalls', 'overalls shirt', 'overalls uniform', 'camouflage', 'heart' } | {'0 feet', '0 ft', '0 inches', '0 length', '0 feet length'} | {'water bottle', 'water', 'wooden bottle', 'wood bottle seed'} |
| W2V | | { 'overalls', 'turkey costume', 'boots', 'vest', 'bow' }<br>– bow | {'3 feet', 'yes', 'ribbon', 'no', 'long'}<br>– long | { 'water', 'river', 'yes', 'pond', 'grass'}<br>– water |
| Semantic | | { 'overalls', 'sweater', 'glasses', 'cape', 'bow'}<br>– sweater | { '3 feet', 'medium length', '5 feet', 'long', 'short'}<br>– 5 feet | { 'pond', 'water', 'cup', 'tea', 'champagne'}<br>– water |
| W2V+Sem | | { 'overalls', 'sweater', 'turkey custome', 'glasses'}<br>– sweater | {'3 feet', 'medium length', 'yes', '5 feet', 'ribbon'}<br>– long | { 'pond', 'water', 'river', 'tea', 'yes'}<br>– river |

Table 2: Sample answer proposals on test2015-standard set. Each cell contains answer proposals in curly brackets along with the final VQA predicted answer following it. Note that the answer proposals are generated by only reading the question, whereas the final VQA answer is also conditioned on the image. The MT system has not been used for training a VQA system, hence, does not have a predicted answer.

model as an answer proposal module in our upcoming VQA experiments.

As Figure 6 shows, the *Semantic* model also comes close to having about 50% of its answers ranked at position 1. This further showcases the strength of our *Semantic* model, suggesting it has higher precision in including relevant answers in the proposal list. Table 2 shows example top-5 ranked answer proposals for various models. As you can see in this table's examples, *W2V* model does not have deep understanding about the SPEECHACT of the questions and sometimes proposes 'yes' or 'no' to wh-questions[12]. Analyzing the proposal lists of our highest recall model, *W2V+Sem*, we have identified three main sources of error. As shown in Table 3, the errors are sometimes due to the missing tokens in multi-word answers or the errors in the human annotations of the VQA dataset[13]. Another source of er-ror is the inherent characteristic of our model, where it only includes the commonsense plausible answers. This results in missing answers such as 'surfing' for the question 'what is unusual about what this dog is doing?'.

In order to further assess the plausibility of the answer proposals, we conducted a human evaluation as follows.

**Human Evaluation on Plausibility**

We conducted human evaluation on a subset of the val set, asking three human judges to rate various systems according to the following prompt.

**Definition 5.1.** *You are shown a question about an image along with a proposed answer. Without seeing the image, do you think the proposed answer can be the actual correct answer to the question?*

For comparison, we also include the top answer list on val set from the SOTA-QA model (example answer list in Table 1). We set the proposal list

---
[12]The Semantic approach has a precision of 99.8% and a recall of 99.2% for detecting YES/NO questions. The error margin is due to occasional parsing errors.

[13]We believe that the human performance of 83.3% on test2015-standar set is partly due to such annotation errors.

|   | **Multi-word prediction error** | **Not including less plausible answers** | **Error in the VQA dataset** |
|---|---|---|---|
| Q. | where is the hand towel? | what is unusual about what this dog is doing? | why does the dog have a cone on his head? |
| A. | on rack | surfing | yes |
| P. | {on wall, on towel rack, next to sink, in air} | { upside down, it is pink, it's rounded, big wheels} | {'pirate hat', 'hat', 'surgery', 'drinking'} |

Table 3: The error analysis of *W2V+Sem* model, where A is the provided correct majority answer and P is the answer proposal list.

cutoff=5 across all models. Table 4 shows the answer proposals generated by our *Semantic* approach on the same questions presented in Table 1. Table 5 shows the human evaluation results. As the results show, our *Semantic* approach comes the closest to our premise of only including truly plausible answers in the answer proposal set. The gap between our answer proposal approaches and the SOTA-QA VQA system is very significant.

## 5.2 Extrinsic Evaluation on VQA

We evaluate our approach for tackling VQA (described in Section 3 and depicted in Figure 2) using various answer proposal modules. We train all the models on trainval2014-train. Table 2 shows the final predicted answers to a few example questions. Table 6 shows the test results on the test2015-standard set. In this table we also include three state-of-the-art models (including the current leader on leaderboard[14]) to be introduced in Section 6. *Toprank* is a baseline which naively predicts the answer to be the highest-ranked answer in *W2V+Sem* proposal list. It is interesting to see that this model can actually predict Yes/No with 70.7% accuracy, which shows the bias of the test dataset. As the results show, *W2V+Sem* is our best performing system, also reflecting on its higher recall. Although our model (which employs a very simple classification module) outperforms some of the state-of-the-art models, it is performing weaker than the best performing systems. Apart from the errors propagated from the Answer Proposal module, outlined in the previous subsection, we hypothesize that QA being constrained by not generating implausible answers can be a challenging task.

---

[14]https://competitions.codalab.org/competitions/6961#results

## 5.3 Revisiting Mutliple-choice VQA

The state-of-the-art accuracy on multiple-choice VQA is higher than open-ended, being a less challenging task. A recent work (Jabri et al., 2016) which studies the biases of the VQA dataset along with the best performing systems shows that their simple binary classification approach (a multi-layer perceptron, which takes in triplet of *(question,image,answer)*) outperforms many of the other complex systems. They suggest two explanations for this observation: (1) the best performing systems are the ones which can best exploit biases in VQA dataset (2) the current VQA models all come short in modeling the problem effectively and reach the same ceiling in accuracy.

Our answer proposal model enables us to shed more light on this matter. We trained the same binary classifier on multiple-choice questions, which achieves 65.2% accuracy on test2015-standard set. Then at test time we swap the multiple-choice list with our pre-generated *W2V+Sem* answer proposal list. We see that the accuracy of the system drops to 49%. This can be partly due to the estimated 8% of the cases in which the correct answer is missing from the answer proposal list. However, as another experiment, we swapped the multiple-choice list with our answer proposal list at train time as well as test time. Then the accuracy of the system increased to 55.1%. This makes the explanation (2) more probable: the current high performing VQA systems are indeed learning the biases of the dataset, where they come short in solving the same task only if provided with a set of more challenging and semantically similar choices[15]. We conclude that training multiple-choice VQA models on a list of plausible choices enables a system to better learn the important features of the question, image, and the answer, however, this will clearly make the multiple-choice VQA task more challenging.

---

[15]The VQA multiple-choice list includes many random and irrelevant options, e.g., the choices in the set {3, no, toothpaste and toothbrush, robin, no idea, red} are all provided for the

| Q. | What is he drinking out of? | What is the dominant vegetable? | What is he mowing? | What is she blowing out? |
|---|---|---|---|---|
| A. | {cup, glass, tub, pond, pool} | {broccoli, carrots, carrot, tomato, pepper, asparagus} | {lawn} | {nose, candles, eggs, windows} |

Table 4: Example answer proposals from *Semantic* model.

| Model | W2V | W2V+Sem | Semantic | SOTA-QA |
|---|---|---|---|---|
| Rating | 75.9 | 77.7 | 94.4 | 66.1 |

Table 5: Normalized average human rating of plausibility of the answer set of various models.

|  | Yes/No | Number | Other | All |
|---|---|---|---|---|
| LSTMImg | 78.9 | 35.2 | 36.4 | 53.7 |
| iBowImg | 76.6 | 35.0 | 42.6 | 55.7 |
| D-NMN | 81.1 | 38.6 | 45.5 | 59.4 |
| MCB | 81.7 | 38.2 | 57.0 | 65.1 |
| Top-rank | 70.7 | 28.6 | 23.4 | 43.4 |
| W2V | 78.8 | 36.6 | 39.7 | 55.4 |
| Semantic | 79.0 | 36.4 | 39.7 | 55.5 |
| W2V+Sem | 79.2 | 36.6 | 40.5 | 55.9 |

Table 6: Comparison of various models on the VQA Real Open-ended task. Results are on the test2015-standard split. Human accuracy on All is 83.3%.

## 6 Related Work

There has been a renewed interest in combining vision and language. VQA is one of the most interesting recent challenges, mainly facilitated by the release of the VQA dataset (Antol et al., 2015), the Toronto COCO-QA (CQA) dataset (Ren et al., 2015), and the Visual7W dataset (Zhu et al., 2016). The VQA dataset is a collection of free-form questions, with both the questions and the set of answers being crowd-sourced. The VQA questions were collected by asking the crowd workers to compose a visually verifiable question which will 'stump a smart robot'. VQA contains 204,721 real images question 'what is on the other side of the train?'. This makes the classification task easier.

and 50,000 abstract images, with various multiple-choice and open-ended questions. Visual7W (Zhu et al., 2016) is another recent dataset, which establishes a grounding link between a textual answer and the regions of the image. This enables answering a question with not only text but also with visual regions. Visual7w contains 327,939 7W multiple-choice QA pairs (but not open-ended questions), including various 'wh' questions. In this paper we base our work on the main VQA challenge dataset[16], specifically open-ended question answering, which is shown to be a more challenging task.

There are various approaches for tackling the task of VQA. The majority of these approaches predict the answer by training a multi-class classifier on image and question features. The classification is performed on the set of unique answers observed in the training set. For this classification there are various neural network architectures combining complex attention mechanisms and memory networks (Lu et al., 2016; Yang et al., 2015). Zhou et al. (Zhu et al., 2016) use deep convolutional features for representing images and averaging word embeddings as question features. The concatenated feature vectors are then fed into a multi-class logistic regression model. Another work (Ma et al., 2016) uses a one-dimensional convolutional network instead of an LSTM encoder for getting the question-level embedding from word-level embeddings.

Another model (*iBOWImg*) (Zhou et al., 2015), is a bag-of-words baseline which concatenates the word features from the question and convolutional features from the image to predict the answer, which shows results competitive with many recent more complex approaches using recurrent neural networks (*LSTMImg*). The Dependency Neural Module Network (*D-NMN*) approach (Andreas et al., 2016) performs dynamic image processing via a compositional network which dynamically restructures it-self using the syntactic parse tree of the question. Fukui et al. (2016) use Multimodal Compact Bilinear pooling (*MCB*) for combining multi-modal (textual and visual) information. They show that multiple MCBs with their architecture with attention achieves the state-of-the-art results on the VQA task. Another recent work (Lu et al., 2016) introduced co-attention of image and question, where they jointly learn a hierarchical attention mechanism based on parsing the question and the image. Recently it has been shown that the attentions generated by neural attention mechanisms are either negatively correlated with where a human looks in the image or if they have positive correlation it is worse than task-independent saliency (Das et al., 2016). Furthermore, many simpler classification approaches (Jabri et al., 2016) are shown to outperform the complex attention architectures that are expected to perform some complex reasoning. This brings up questions regarding the effectiveness of the current complex approaches to VQA and further reveals the biases of the VQA multiple-choice question set (Jabri et al., 2016).

---
[16]http://visualqa.org/download.html

There are also approaches that look at the problem not as a classification task but as a generative one. Notable among the generative approaches, Malinowski et al. (2014) generate the answer using an LSTM which is conditioned on the deep convolutional image features and the question. Although generative models are a promising way to generalize the production of unseen answers during training, the earlier work showed that joint learning of the encoding and decoding models from the VQA datasets has not been successful. More importantly, it is not clear how to automatically evaluate the novel generated text.

Our deep binary classification module (Section 3) is more closely related to Shih et al. (Shih et al., 2016) and Jabri et al. (Jabri et al., 2016) which also take the answer as an input variable to a classifier that then assigns a probability to the *(question, image, answer)* triplet as a whole. While we use deep convolutional features as our visual features, Shih et al. use a more complex image processing module where they select image regions for answering. However, none of the earlier work propose an effective approach for open-ended question answering, where the 'answer' in the *(question, image, answer)* triplet is not given. Our work introduces a novel perspective for tackling open-ended VQA questions which has not been explored by any of the previous work: answer proposals. We provide various methods for proposing plausible answers, where the semantic-driven approach outperforms others.

The impact of semantics as opposed to surface n-gram wording of textual content has also been studied in SPICE captioning evaluation (Anderson et al., 2016). SPICE emphasizes on the importance of semantic propositional content of captions, as captured by dependency parse trees, which correlates well with how human evaluates captions. We also note the work of (Xu et al., 2015) that while focused on caption generation and retrieval tasks for video using a joint language and vision model, proposed a compositional semantics language model that enforced semantic compatibility between essential concepts, similar to our goal of using question semantics to constrain our answer proposals.

# 7 Conclusion

We introduced the novel task of proposing plausible answers for a given open-ended question, where a system should generate a ranked list of plausible answers given a question. We use the VQA task as a multimodal test framework for training and testing answer proposal models. We provide various answer proposal models, ranging from vector-based to deep semantic ones. We show that our best performing model which combines our two retrieval models achieves a high recall. Furthermore, we show that our Semantic Graph Matching approach generates truly plausible answers, unlike the state-of-the-art models. Our full VQA model outperforms some other solutions to VQA, however, performs weaker than the current best performing systems. We hypothesize that answering questions with the condition of generating only plausible answers can be more challenging that only answering questions. Our next step is to employ better answer proposal models, possibly by injecting external world knowledge from other resources. Although we have mainly experimented with the VQA task, similar Answer Proposal models can be potentially used in other QA tasks.